\documentclass[11pt,a4paper]{article}
\usepackage[hyperref]{acl2020}
\usepackage{times}
\usepackage{latexsym}
\usepackage{amsmath}
\usepackage{booktabs}
\usepackage{subfig}
\usepackage{graphicx}
\usepackage{xspace}
\usepackage{latexsym}
\usepackage{multirow}
\usepackage{amsmath}
\usepackage{colortbl}
\DeclareMathOperator*{\argmax}{argmax}

\newenvironment{itemize*}%
 {\leftmargini=10pt\begin{itemize}%
  \setlength{\itemsep}{0pt}%
  \setlength{\parskip}{0pt}%
  }%
 {\end{itemize}}
\newenvironment{enumerate*}%
 {\begin{enumerate}%
  \setlength{\itemsep}{0pt}%
  \setlength{\parskip}{0pt}}%
 {\end{enumerate}}

\definecolor{myblue}{rgb}{0.9, 0.1, 0.94}
\definecolor{mygreen}{rgb}{0.64, 0.56, 0.88}
\definecolor{myyellow}{rgb}{0.98, 0.94, 0.75}
\definecolor{mygreen}{rgb}{0.68, 0.9, 0.6}
\definecolor{myorange}{rgb}{1.0, 0.49, 0.0}	
\definecolor{lightgray}{rgb}{0.83, 0.83, 0.83}
\definecolor{dimgray}{rgb}{0.41, 0.41, 0.41}
\definecolor{mygrey}{rgb}{0.92, 0.92, 0.92}

\usepackage{microtype}
\usepackage{afterpage}
\aclfinalcopy 

\author{Jinlan Fu \thanks{\ \ This work is done when Jinlan visited CMU remotely.} \\
  Fudan University \\
  \texttt{fujl16@fudan.edu.cn} \\
  \And
  Xuanjing Huang  \\
  Fudan University \\
\texttt{xjhuang@fudan.edu.cn}
  \And
  Pengfei Liu \thanks{\ \  Corresponding author.} \\
  Carnegie Mellon University \\
\texttt{pliu3@cs.cmu.edu} 
  }

\title{\textsc{SpanNer}: Named Entity \textit{Re}-/Recognition as Span Prediction}

\date{}

\begin{document}
\maketitle
\begin{abstract}
    Recent years have seen the paradigm shift of Named Entity Recognition (NER) systems from \textit{sequence labeling} to \textit{span prediction}.
    Despite its preliminary effectiveness, the span prediction model's architectural bias has not been fully understood.
   In this paper, we first investigate the strengths and weaknesses when the span prediction model is used for \textit{named entity recognition} compared with the sequence labeling framework and how to further improve it, which motivates us to make complementary advantages of systems based on different paradigms.
    We then reveal that span prediction, simultaneously, can serve as a system combiner to \textit{re-recognize} named entities  from different systems' outputs.
    We experimentally implement 154 systems on 11 datasets, covering three languages, comprehensive results show the effectiveness of span prediction models that both serve as base NER systems and system combiners.
    We make all code and datasets available:
    \url{https://github.com/neulab/spanner}, 
    as well as an online system demo: \url{http://spanner.sh}.
    Our model also has been deployed into the \textsc{ExplainaBoard} \cite{liu2021explainaboard} platform, which allows users to flexibly perform the system combination of top-scoring systems in an interactive way: \url{http://explainaboard.nlpedia.ai/leaderboard/task-ner/}.
  
\end{abstract}

\section{Introduction}

The rapid evolution of neural architectures \cite{kalchbrenner-etal-2014-convolutional,kim-2014-convolutional,hochreiter1997long} and large pre-trained models \cite{devlin2019bert,lewis-etal-2020-bart} not only drive the state-of-the-art performance of many NLP tasks \cite{devlin2019bert,liu-lapata-2019-text} to a new level but also change the way how researchers formulate the task.
For example, recent years have seen frequent paradigm shifts for the task of named entity recognition (NER)  from \textit{token-level tagging}, which conceptualize NER as a sequence labeling (\textsc{SeqLab}) task \cite{chiu2015named,huang2015bidirectional,ma2016end,lample2016neural,akbik2018contextual,peters2018deep,devlin2018bert,xia2019multi,luo2020hierarchical,lin2020triggerner,fu2021larger}, to \textit{span-level prediction} (\textsc{SpanNer})  \cite{li-etal-2020-unified,mengge2020coarse,jiang-etal-2020-generalizing,ouchi2020instance,yu-etal-2020-named}, which regards NER either as question answering \cite{li-etal-2020-unified,mengge2020coarse}, span classification~\cite{jiang-etal-2020-generalizing,ouchi2020instance,yamada-etal-2020-luke}, and dependency parsing tasks \cite{yu-etal-2020-named}.

\begin{figure}
    \centering
    \includegraphics[width=0.75\linewidth]{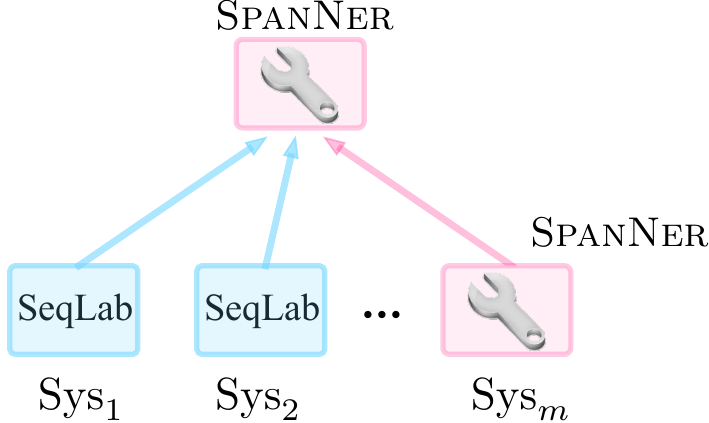}
    \vspace{-6pt}
    \caption{\textbf{ONE} span prediction model (\textsc{SpanNER}) finishes \textbf{TWO} things: (1) named entity recognition (2) combination of different NER systems.
    }
    \label{fig:intro_nerr}
\end{figure}

However, despite the success of span prediction-based systems, as a relatively newly-explored framework, the understanding of its architectural bias has not been fully understood so far. For example, what are the complementary advantages compared with \textsc{SeqLab} frameworks and how to make full use of them?
Motivated by this, in this paper, we make two scientific contributions.

We first investigate \textbf{what strengths and weaknesses are when NER is conceptualized as a span prediction task}.
To achieve this goal, we perform a fine-grained evaluation of \textsc{SpanNer} systems against \textsc{SeqLab} systems and find there are clear complementary advantages between these two frameworks. For example, \textsc{SeqLab}-based models are better at dealing with those entities that are long and with low label consistency. By contrast, \textsc{SpanNer} systems do better in sentences with more Out-of-Vocabulary (OOV) words and entities with medium length (\S\ref{exp-II}).

Secondly, we \textbf{reveal the unique advantage brought by the architectural bias of the span prediction framework}: \textit{it can not only be used as a base system for named entity recognition but also serve as a meta-system to combine multiple NER  systems' outputs}. In other words, the span prediction model play two roles showing in Fig.~\ref{fig:intro_nerr}: (i) as a base NER system; and (ii) as a system combiner of multiple base systems.  
We claim that compared with traditional ensemble learning of the NER task, \textsc{SpanNer} combiners are advantageous in the following aspects:

\begin{enumerate*}
    \item 
    Most of the existing NER combiners rely on heavy feature engineering and external knowledge~\cite{florian2003named,wu2003stacked,saha2013combining}. Instead, the \textsc{SpanNer} models we proposed for system combination train in an end-to-end fashion.
    \item Combining complementarities of different paradigms: most previous works perform NER system combination solely focusing on the sequence labeling framework. It is still an understudied topic how systems from different frameworks help each other.
    \item No extra training overhead and flexibility of use: Existing ensemble learning algorithms are expensive, which usually need to collect training samples by k-fold cross-validation for system combiner~\cite{speck2014ensemble}, reducing their practicality.
    \item Connecting two separated training processes: previously, the optimization of base NER systems and ensemble learning for combiner are two independent processes. Our work builds their connection and the same set of parameters shared over these two processes.
\end{enumerate*}

Experimentally, we first implement 154 systems on 11 datasets, on which we comprehensively evaluate the effectiveness of our proposed span prediction-based system combiner. Empirical results show its superior performance against several typical ensemble learning algorithms.

Lastly, \textbf{we make an engineering contribution that benefits from the practicality of our proposed methods}.
Specifically, we developed an online demo system based on our proposed method, and 
integrate it into the NER Leaderboard, which is very convenient for researchers to find the complementarities among different combinations of systems, and search for a new state-of-the-art system.

\section{Preliminaries}

\subsection{Task}
NER is frequently formulated as a sequence labeling (\textsc{SeqLab}) problem \citep{chiu2015named,huang2015bidirectional,ma2016end,lample2016neural}, where $X = \{x_1,x_2,\ldots, x_T\}$ is an input sequence and $Y = \{y_1,y_2,\ldots,y_T\}$ is the output label (e.g., ``B-PER'', ``I-LOC'', ``O'') sequence.
The goal of this task is to accurately predict entities by assigning output label $y_t$ for each token $x_t$. We take the F1-score\footnote{\url{http://www.clips.uantwerpen.be/conll2000/chunking/conlleval.txt}} as the evaluation metric for the NER task.

\subsection{Datasets}
To make a comprehensive evaluation, in this paper, we use multiple NER datasets that cover different domains and languages.

\noindent\textbf{CoNLL-2003} \footnote{\url{https://www.clips.uantwerpen.be/conll2003/ner/}} \citep{sang2003introduction}
covers two different languages: English and German.
Here, we only consider the  English (\texttt{EN}) dataset collected from the Reuters Corpus.

\noindent\textbf{CoNLL-2002} \footnote{\url{https://www.clips.uantwerpen.be/conll2002/ner/}} \citep{sang2002introduction} contains annotated corpus in Dutch (\texttt{NL}) collected from De Morgen news, and Spanish (\texttt{ES}) collected from Spanish EFE News Agency. We evaluate both languages.   

\noindent\textbf{OntoNotes 5.0} \footnote{\url{https://catalog.ldc.upenn.edu/LDC2013T19}} \citep{weischedel2013ontonotes}  is a large corpus consisting of three different languages: English, Chinese, and Arabic, involving six genres: newswire (\texttt{NW}), broadcast news (\texttt{BN}), broadcast conversation (\texttt{BC}), magazine (\texttt{MZ}),   web data (\texttt{WB}), and telephone conversation (\texttt{TC}). Following previous works \cite{durrett2014joint,ghaddar2018robust}, we utilize different domains in English to test the robustness of proposed models.

\noindent\textbf{WNUT-2016 \footnote{\url{http://noisy-text.github.io/2016/ner-shared-task.html}} and WNUT-2017 \footnote{\url{http://noisy-text.github.io/2017/emerging-rare-entities.html}} } \citep{strauss2016results,derczynski2017results}  are social media data from Twitter, which were public as a shared task at  WNUT-2016 (\texttt{W16}) and WNUT-2017 (\texttt{W17}).

\section{Span Prediction for NE Recognition}
Although this is not the first work that formulates NER as a span prediction problem \cite{jiang-etal-2020-generalizing,ouchi2020instance,yu-etal-2020-named,li-etal-2020-unified,mengge2020coarse},  we contribute by  (1) exploring how  different design choices influence the performance of \textsc{SpanNer} and (2) interpreting complementary strengths between \textsc{SeqLab} and \textsc{SpanNer} with different design choices.
In what follows, we first detail span prediction-based NER systems with the vanilla configuration and  proposed advanced featurization.

\subsection{\textsc{SpanNer} as NER System}
Overall, the span prediction-based framework for NER consists of three major modules: token representation layer, span representation layer, and span prediction layer.

\subsubsection{Token Representation Layer}
\label{sec:token_rep}
Given a sentence $X=\{ x_{1}, \cdots, x_{n}\}$ with $n$ tokens, the token representation $\mathbf{h}_i$ is as follows:
\begin{align}
    \mathbf{u}_1, \cdots, \mathbf{u}_n &= \textsc{Emb}(x_1, \cdots, x_n), \\
     \mathbf{h}_1, \cdots, \mathbf{h}_n &= \textsc{BiLSTM}( \mathbf{u}_1, \cdots, \mathbf{u}_n), 
\end{align}
\noindent
where $\textsc{Emb}(\cdot)$ is the pre-trained embeddings, such as non-contextualized embeddings GloVe~\citep{pennington2014glove} or contextualized pre-trained embeddings BERT~\citep{devlin2018bert}. $\textsc{BiLSTM}$ is the bidirectional LSTM~\cite{hochreiter1997long}.

\subsubsection{Span Representation Layer}
First, we enumerate all the possible $m$ spans $S=\{s_1,\cdots, s_i, \cdots, s_m\}$ for sentence $X=\{x_{1}, \cdots, x_{n}\}$ and then re-assign a label $y\in \mathcal{Y}$ for each span $s$. 
For example, for sentence: ``London$_1$ is$_2$ beautiful$_3$'', the possible span's (start, end) indices are $\{(1,1), (2,2),(3,3),(1,2),(2,3),(1,3)\}$, and the labels of these spans are all ``O'' except $(1,1)$ (London) is ``LOC''.
We use $b_i$ and $e_i$ to denote the start- and end- index of the span $s_i$, respectively, and $1 \leq b_i \leq e_i \leq n$. Then each span can be represented as $s_i=\{x_{b_i}, x_{b_i+1},\cdots,x_{e_i}\}$. 
The vectorial representation of each span could be calculated based on the following parts:

\noindent \textbf{Boundary embedding:} span representation is calculated by the concatenation of the start and end tokens' representations $\mathbf{z}_i^b =[\mathbf{h}_{b_{i}};\mathbf{h}_{e_{i}}]$

\noindent \textbf{Span length embedding:} we additionally featurize each span representation by introducing its length embedding $\mathbf{z}_i^{l}$, which can be obtained by a learnable look-up table.

The final representation of each span $s_i$ can be obtained as: $\mathbf{s}_i =[\mathbf{z}_i^b;\mathbf{z}_i^l] $.

\subsubsection{Span Prediction Layer}
The span representations $\mathbf{s}_i$ are fed into a softmax function to get the probability w.r.t label $y$. 
\begin{align}
\text{P}(y | s_i) & = \frac{\mathrm{score}(\mathbf{s}_i, \mathbf{y})}{\displaystyle \sum_{y' \in \mathcal{Y}} \mathrm{score}(\mathbf{s}_i, \mathbf{y}')}, 
\end{align}
\noindent
where $\mathrm{score}(\cdot)$ is a function that measures the compatibility between a specified label and a span:
\begin{align}
    \mathrm{score}(\mathbf{s}_i, \mathbf{y}_k) = \exp({\mathbf{s}_i^T\mathbf{y}_k}),  \label{eq:score}
\end{align}

\noindent
where $\mathbf{s}_i$ denotes the span representation and $\mathbf{y}_k$ is a learnable representation of the class $k$.

\begin{table*}[!htb]
  \centering \footnotesize
    \begin{tabular}{lccccccccccc}
    \toprule
    \multirow{2}[4]{*}{Model} & \multicolumn{3}{c}{CoNLL} & \multicolumn{6}{c}{OntoNotes 5.0}             & \multicolumn{2}{c}{WNUT} \\
\cmidrule(lr){2-4} \cmidrule(lr){5-10}  \cmidrule(lr){11-12} 
& EN    & ES    & NL    & BN    & BC    & MZ    & WB    & NW    & TC    & W16 & W17 \\
    \midrule
    generic  & 91.57 & 84.58 & 88.79 & 89.66 & 82.24 & 85.42 & 67.92 & 90.84 & 66.67 & 55.70     & 52.05 \\
    + decode$*$ & 91.89  & 85.34 & 89.56 & 90.55 & 82.79 & 86.62 & 68.01 & 91.01 & 68.54 & 55.72     & 52.59 \\
    + length$*$ & 92.22 & 84.82 & 89.81 & 90.60  & 83.01 & 86.28 & 66.69 & 91.31 & 68.96 & 55.78     & 52.58 \\
    + both$*$ & 92.28 & 87.54 & 91.04 & 90.93 & 83.22 & 87.03 & 68.58 & 91.59 & 69.91 & 56.27     & 52.97 \\
    \bottomrule
    \end{tabular}
    \vspace{-6pt}
      \caption{The results of the span prediction model with different features. $*$ denotes that the model's performance is significantly better than the generic setting ($p<0.01$).
      } 
  \label{tab:spanseqlab}
\end{table*}

\paragraph{Heuristic Decoding}
Regarding the flat NER task without nested entities, we present a heuristic decoding method to avoid the prediction of overlapped spans. Specifically, for those overlapped spans, we keep the span with the highest prediction probability and drop the others.

\subsection{Exp-I: Effectiveness of Model Variants}

\paragraph{Setup}
To explore how different mechanisms influence the performance of span prediction models, We design four specific model variants (i) generic \textsc{SpanNer}: only using boundary embedding (ii) boundary embedding + span length embedding, (iii) boundary embedding + heuristic decoding, (iv) heuristic decoding + (ii).

\renewcommand\tabcolsep{2.2pt}
\begin{table*}[htb]
  \centering \footnotesize
    \begin{tabular}{rrrrrrrrrrr rrrrrrrrrrr}
    \toprule
    \multicolumn{11}{c}{Generic (\textsc{SpanNer}), F1: 91.57} & \multicolumn{11}{c}{Generic+decode, F1: 91.89}\\
    \cmidrule(lr){1-11}\cmidrule(lr){12-22}
    \multicolumn{4}{c}{eCon}      & \multicolumn{2}{c}{sLen} & \multicolumn{2}{c}{eLen} & \multicolumn{3}{c}{oDen} 
    & \multicolumn{4}{c}{eCon}      & \multicolumn{2}{c}{sLen} & \multicolumn{2}{c}{eLen} & \multicolumn{3}{c}{oDen}\\
    \cmidrule(lr){1-11}\cmidrule(lr){12-22}
           \multicolumn{4}{c}{\includegraphics[scale=0.17]{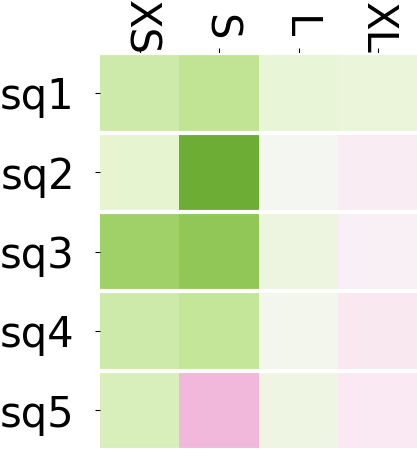}}  
        & \multicolumn{2}{c}{\includegraphics[scale=0.17]{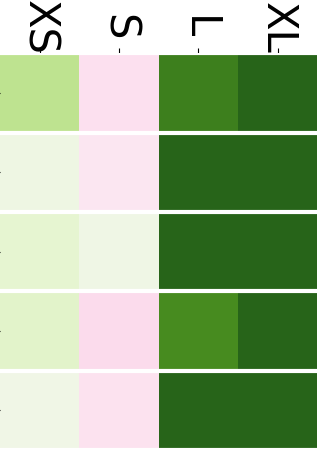}}                 
        & \multicolumn{2}{c}{\includegraphics[scale=0.17]{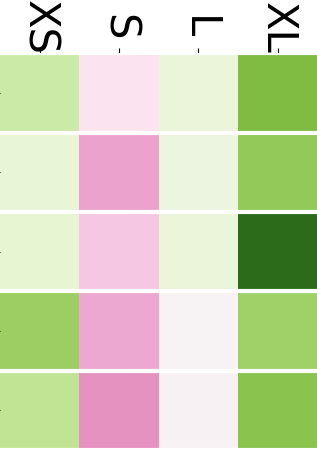}} 
        & \multicolumn{3}{c}{\includegraphics[scale=0.17]{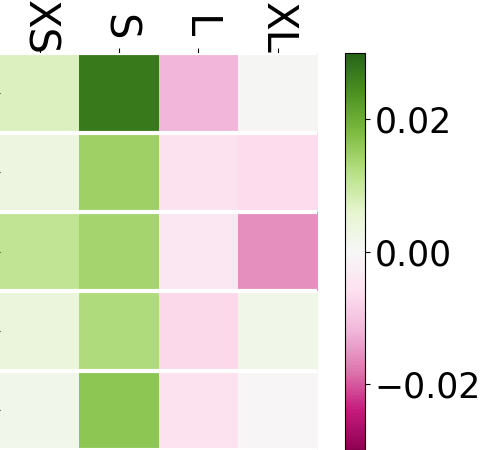}} 
        & \multicolumn{4}{c}{\includegraphics[scale=0.17]{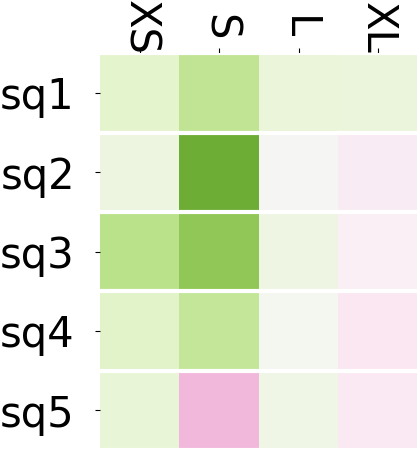}}  
        & \multicolumn{2}{c}{\includegraphics[scale=0.17]{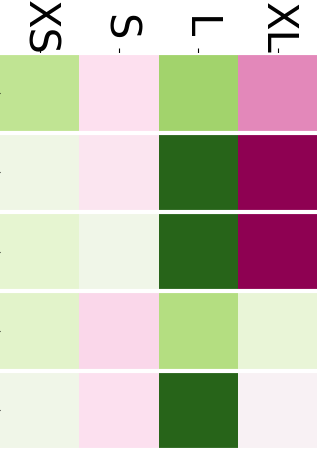}}                 
        & \multicolumn{2}{c}{\includegraphics[scale=0.17]{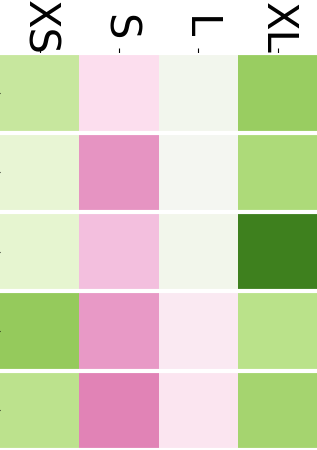}} 
        & \multicolumn{3}{c}{\includegraphics[scale=0.17]{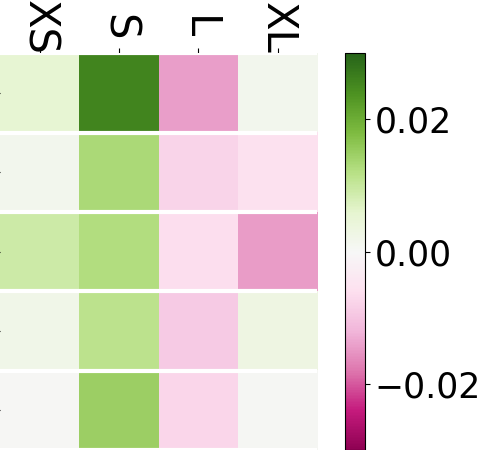}} 
        \\
        \midrule
        \midrule
        \multicolumn{11}{c}{Generic+length, F1: 92.22} & \multicolumn{11}{c}{Generic+length+decode, F1: 92.28}\\
    \cmidrule(lr){1-11}\cmidrule(lr){12-22}
    \multicolumn{4}{c}{eCon}      & \multicolumn{2}{c}{sLen} & \multicolumn{2}{c}{eLen} & \multicolumn{3}{c}{oDen} 
    & \multicolumn{4}{c}{eCon}      & \multicolumn{2}{c}{sLen} & \multicolumn{2}{c}{eLen} & \multicolumn{3}{c}{oDen}\\
    \cmidrule(lr){1-11}\cmidrule(lr){12-22}
          \multicolumn{4}{c}{\includegraphics[scale=0.17]{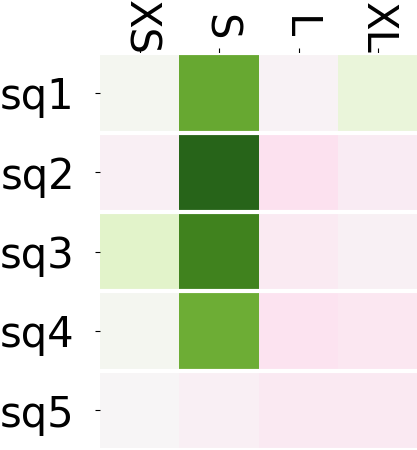}}  
        & \multicolumn{2}{c}{\includegraphics[scale=0.17]{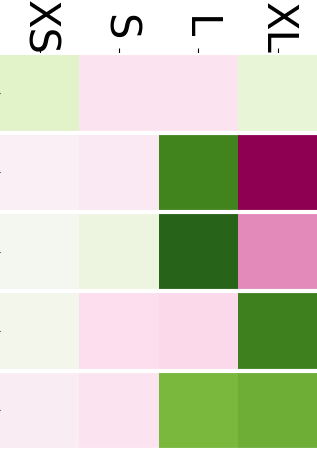}}                 
        & \multicolumn{2}{c}{\includegraphics[scale=0.17]{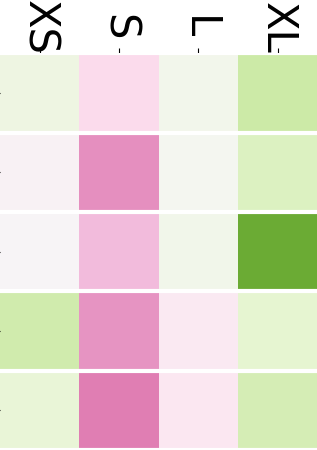}} 
        & \multicolumn{3}{c}{\includegraphics[scale=0.17]{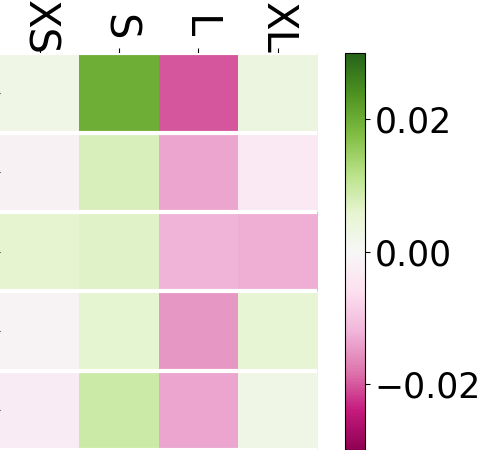}} 
        & \multicolumn{4}{c}{\includegraphics[scale=0.17]{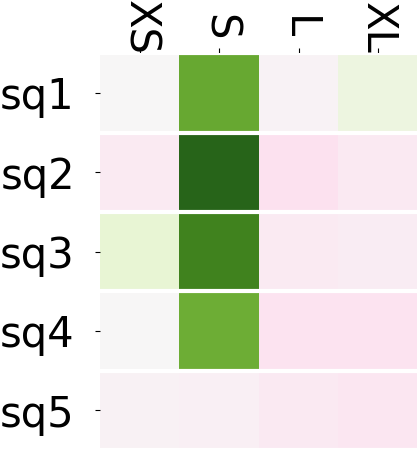}}  
        & \multicolumn{2}{c}{\includegraphics[scale=0.17]{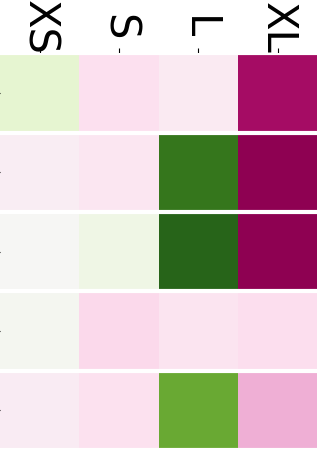}}                 
        & \multicolumn{2}{c}{\includegraphics[scale=0.17]{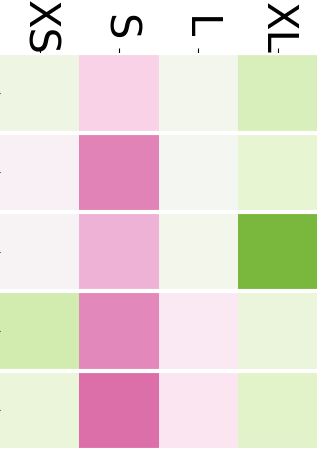}} 
        & \multicolumn{3}{c}{\includegraphics[scale=0.17]{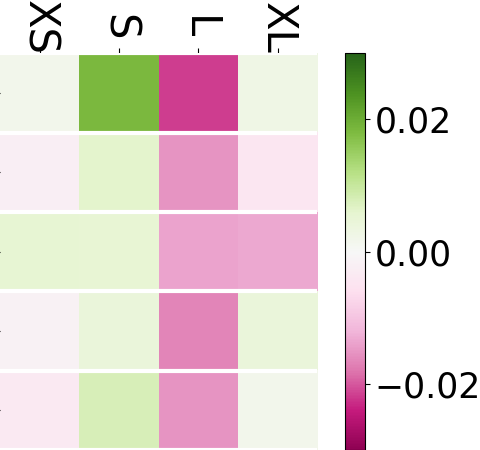}} 
        \\
    \bottomrule
    \end{tabular}
    \vspace{-6pt}
      \caption{Performance heatmap of pair-wise system diagnosis. $sq_i$ represents different \textsc{SeqLab} systems. Each value in heatmap entry $(i,j)$ represents the performance gap between \textsc{SeqLab} and \textsc{SpanNer} ($F1_{\text{sq}} - F1_{\text{span}}$) on $j$-th bucket. \textbf{The green area indicates \textsc{SeqLab} performs better while the red area implies \textsc{SpanNer} is better.} \texttt{eCon}, \texttt{sLen}, \texttt{eLen}, and \texttt{oDen} represent different attributes. }
  \label{tab:exp-II-heatmap}
\end{table*}

\paragraph{Results}
As shown in Tab.~\ref{tab:spanseqlab}, we can observe that:
(i) heuristic decoding is an effective method that can boost the generic model's performance over all the datasets.
(ii) span length feature works most of the time. The performances on $10$ of the $11$ datasets have improved against the generic model.
(iii) By combining two mechanisms together, significant improvements were achieved on all datasets.

\subsection{Exp-II: Analysis of Complementarity} \label{exp-II}
The holistic results in Tab.~\ref{tab:spanseqlab} make it hard for us to interpret the relative advantages of NER systems with different structural biases.
To address this problem, we follow the interpretable evaluation idea \cite{fu-etal-2020-interpretable,fu-etal-2020-rethinkcws} that proposes to breakdown the holistic performance into different buckets from different perspectives and use a \textit{performance heatmap} to illustrate relative advantages between two systems, i.e., system-pair diagnosis.

\paragraph{Setup}
As a comparison, we replicate five top-scoring \textsc{SeqLab}-based NER systems, which are 
$sq1: 92.41$, 
$sq2: 92.01$,
$sq3: 92.46$,
$sq4: 92.11$,
$sq5: 91.99$.
Notably, to make a fair comparison, all five \textsc{SeqLab}s are with closed performance comparing to the above \textsc{SpanNer}s.
Although we will detail configurations of these systems later (to reduce content redundancy) in \S\ref{sec:basesystem} Tab.~\ref{tab:seqmodels-res} , it would not influence our analysis in this section.

Regarding interpretable evaluation, we choose the \texttt{CoNLL-2003 (EN)} dataset as a case study and breakdown the holistic performance into four groups based on different attributes.
Specifically, given an entity  ${e}$ that belongs to a sentence $S$, the following attribute feature functions can be defined:

\vspace{-5pt}
\begin{itemize*}
    \item $\phi_{\texttt{eLen}} = \mathrm{len}(e)$: \textit{entity length}
    \item $\phi_{\texttt{sLen}} =  \mathrm{len}(S)$: \textit{sentence length}
    \vspace{3pt}
    \item  $\phi_{\texttt{oDen}} = \frac{|\text{OOVs}|}{\mathrm{len}(S)}$: \textit{density of OOVs}
    \vspace{3pt}
    \item  $\phi_{\texttt{eCon}} = \frac{|\{\varepsilon| \mathrm{label}(\varepsilon) = \mathrm{label}(e) , \forall \varepsilon \in \mathcal{E}\}|}{|\mathcal{E}|}$
    : \textit{entity label consistency}
\end{itemize*}
\vspace{-5pt}

\noindent where $\mathrm{len}(\cdot)$ counts the number of words, $\mathrm{label}(e)$ gets the label for span $e$, $\mathcal{E}$ denotes all spans in the training set.  $|\text{OOVs}|$ is the number of OOV words in the sentence.

We additionally use a training set dependent attribute: entity label consistency (\texttt{eCon}), which measures how consistently a particular entity is labeled with a particular label. For example, if an entity with the label ``LOC'' has a higher \texttt{eCon}, it means that the entity is frequently labeled as ``LOC'' in the training set. 
Based on the attribute value of entities, we partition test entities into four buckets: extra-small (XS), small (S), large (L), and extra-large (XL).\footnote{we show detailed bucket intervals in the appendix}. For each bucket, we calculate a bucket-wise F1.

\paragraph{Analysis}
As shown in Tab.~\ref{tab:exp-II-heatmap}, the green area indicates \textsc{SeqLab} performs better while the red area implies the span model is better.
We observe that:

\noindent
(1) The generic \textsc{SpanNer} shows clear complementary advantages with \textsc{SeqLab}-based systems.
Specifically, almost all \textsc{SeqLab}-based models outperform generic \textsc{SpanNer} when (i) entities are long and with lower label consistency (ii) sentences are long and with fewer OOV words. 
By contrast, \textsc{SpanNer} is better at dealing with entities locating on sentences with more OOV words and entities with medium length.

\noindent (2) By introducing heuristic decoding and span length features, \textsc{SpanNer}s do slightly better in long sentences and long entities, but are still underperforming on entities with lower label consistency.

The complementary advantages presented by \textsc{SeqLab}s and \textsc{SpanNer}s motivate us to search for an effective framework to utilize them.

\section{Span Prediction for NE Re-recognition}

The development of ensemble learning for NER systems, so far, lags behind the architectural evolution of the NER task.
Based on our evidence from \S\ref{exp-II}, we propose a new ensemble learning framework for NER systems.

\paragraph{\textsc{SpanNer} as System Combiner}

\renewcommand\tabcolsep{1.8pt}
\renewcommand{\arraystretch}{1.1}
\begin{table*}[htb]
  \centering \footnotesize
    \begin{tabular}{lccccccccccccccccccccc}
    \toprule
    \multicolumn{1}{c}{\multirow{2}[4]{*}{Models}} & \multicolumn{5}{c}{Char/Sub.}      & \multicolumn{3}{c}{Word} & \multicolumn{2}{c}{Sent.}  & \multicolumn{3}{c}{CoNLL} & \multicolumn{6}{c}{OntoNotes 5.0}                     & \multicolumn{2}{c}{WNUT} \\
    \cmidrule(lr){2-6}\cmidrule(lr){7-9}\cmidrule(lr){10-11} \cmidrule(lr){12-14} \cmidrule(lr){15-20}  \cmidrule(lr){21-22}          & \rotatebox{90}{none}  & \rotatebox{90}{cnn}   & \rotatebox{90}{elmo}  & \rotatebox{90}{flair} & \rotatebox{90}{bert}  & \rotatebox{90}{none}  & \rotatebox{90}{rand}  & \rotatebox{90}{glove} & \rotatebox{90}{lstm}  & \rotatebox{90}{cnn}    & EN    & ES    & NL       & BN    & BC    & MZ    & WB    & NW    & TC    & 
   W16& W17 \\
    \midrule
    sq0 &       &       &       & $\surd$     &       &       &       & $\surd$     & $\surd$     &      & 93.02 & 87.87 & 87.76     & 89.43 & 78.17 & 88.24 & 67.19 & 90.11 & 66.57 & 52.07 & 44.75 \\
     sq1 &       &       &       & $\surd$     &       & $\surd$    &       &       & $\surd$     &       & 92.41 & 88.11 & 87.71    & 89.03 & 79.55 & 87.13 & 67.78 & 90.23 & 65.58 & 52.22 & 43.57 \\
    sq2 &       &       &       &       & $\surd$     &       &       & $\surd$     & $\surd$     &       & 92.01 & 88.81 & 91.73    & 90.70  & 81.55 & 88.02 & 62.14 & 90.08 & 71.07 & 50.18 & 45.23 \\
      sq3 &       &       &       &       & $\surd$     & $\surd$     &       &       & $\surd$     &        & 92.46 & 88.00    & 91.34   & 90.53 & 80.11 & 88.87 & 62.90  & 90.77 & 71.01 & 49.87 & 46.47 \\
     sq4 &       &       & $\surd$    &       &       &       &       & $\surd$     & $\surd$    &      & 92.11 & - & -        & 89.33 & 78.28 & 85.84 & 62.62 & 90.10  & 64.62 & 50.22 & 48.91 \\
     sq5 &       &       & $\surd$     &       &       & $\surd$     &       &       & $\surd$     &      & 91.99 & - & -      & 89.21 & 79.32 & 84.64 & 61.69 & 90.44 & 65.57 & 49.86 & 47.35 \\
    sq6 &       & $\surd$    &       &       &       &       &       & $\surd$      & $\surd$     &        & 90.88 & 82.33 & 82.23     & 86.84 & 75.10  & 86.61 & 62.61 & 88.31 & 64.36 & 42.04 & 36.41 \\
    sq7 &       & $\surd$     &       &       &       &       &       & $\surd$     &       & $\surd$       & 89.71 & 80.01 & 80.70      & 86.18 & 74.63 & 86.55 & 49.85 & 86.87 & 56.16 & 39.40  & 33.72 \\
    sq8 &       & $\surd$     &       &       &       &       & $\surd$     &       & $\surd$     &         & 83.03 & 79.44 & 75.44     & 83.87 & 69.81 & 82.20  & 51.35 & 86.03 & 51.83 & 20.68 & 18.77 \\
    sq9 & $\surd$     &       &       &       &       &       & $\surd$     &       & $\surd$     &        & 78.49 & 70.66 & 64.78     & 81.05 & 66.42 & 75.34 & 48.91 & 85.73 & 46.84 & 17.24 & 18.39 \\
    \midrule
     \multicolumn{22}{l}{\textsc{SpanNer}} \\
     + generic (sp1)  &     &       &       &       &       &       &     &       &      &   & 91.57 & 84.58 & 88.79 & 89.66 & 82.24 & 85.42 & 67.92 & 90.84 & 66.67 & 55.70     & 52.05 \\
    + both (sp2) &     &       &       &       &       &       &     &       &      &    &  92.28 & 87.54 & 91.04 & 90.93 & 83.22 & 87.03 & 68.58 & 91.59 & 69.91 & 56.27     & 52.97 \\
    \bottomrule
    \end{tabular}
    \vspace{-6pt}
       \caption{The holistic performance of the $12$ base models on $11$ datasets. 
       ``Sub'' and ``sent.'' denotes the subword embedding and sentence encoder, respectively. All the ten sequence labeling models use the CRF as the decoder. $\surd$ indicates the embedding/structure is utilized in the current \textsc{SeqLab} system. For example,
       ``sq0'' denotes a model that uses Flair, GloVe, LSTM, and CRF as the character-, word-level embedding, sentence-level encoder, and decoder, respectively. ``--'' indicates not applicable\protect.\footnotemark{}}
  \label{tab:seqmodels-res}
\end{table*}
\afterpage{
\footnotetext{Since the lack of an official EMLo language model in Spanish and Dutch, we do not implement these base models.}
}

\begin{figure}[!ht]
    \centering
    \includegraphics[width=1.0\linewidth]{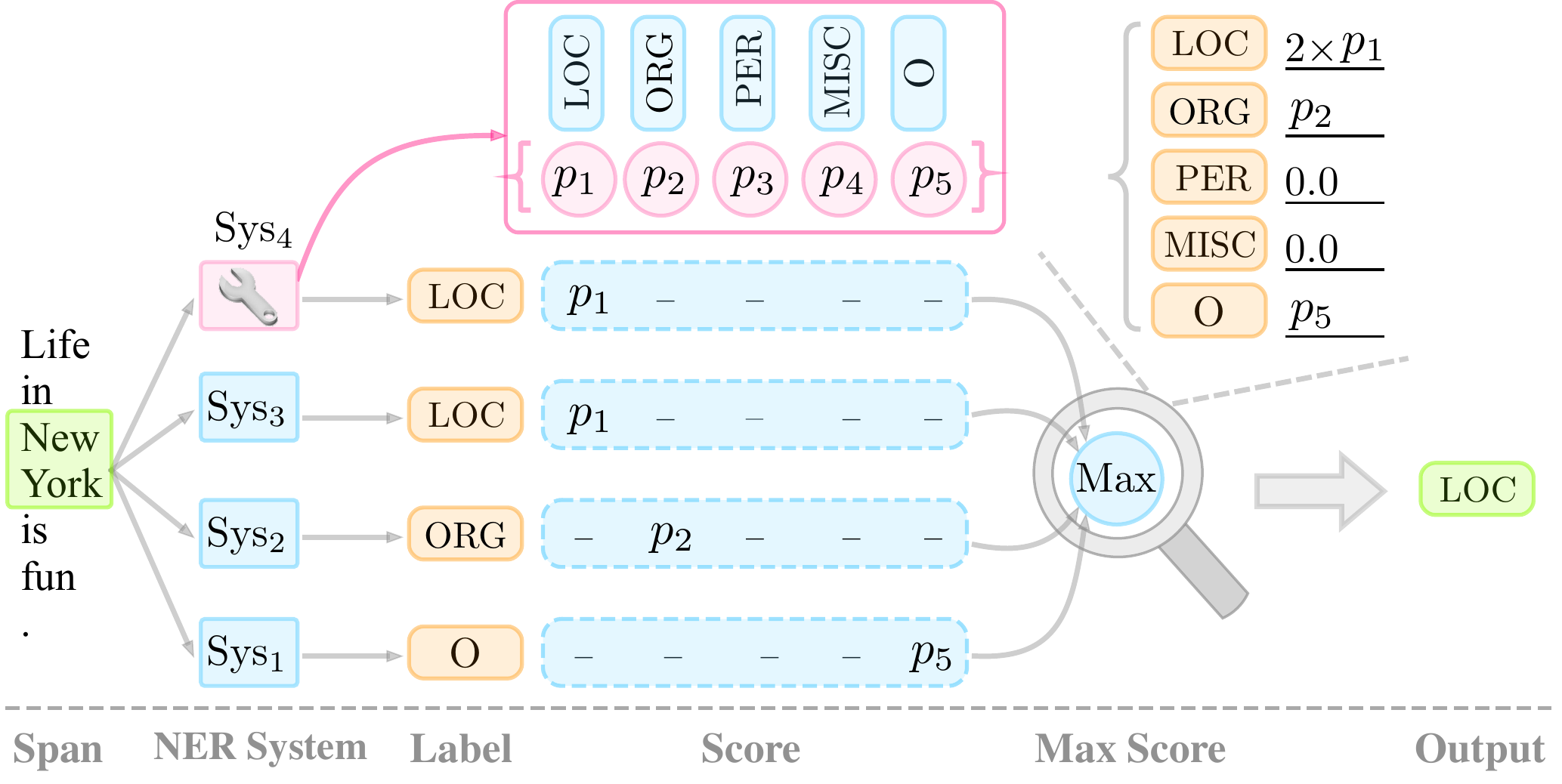}
    \vspace{-6pt}
    \caption{The framework of span prediction system (\textsc{SpanNer}) as system combiner.  $p_i$ is calculated by function $\mathrm{score}(\cdot)$, as defined in Eq.\ref{eq:score}. $\mathrm{score}(\cdot)$ takes a span and predicted label as input and output a matching score. For example, $\mathrm{score}(\texttt{New York}, \texttt{LOC}) = 0.5$ suggests that span ``\texttt{New York}'' matches ``\texttt{LOC}'' with the 
   score of 0.5.  
    }
    \label{fig:sp-combiner}
\end{figure}

The basic idea is that each span prediction NER (\textsc{SpanNer}) system itself can also conceptualize as a system combiner to re-recognize named entities from different systems' outputs.
Specifically, Fig.~\ref{fig:sp-combiner} illustrates the general workflow. 
Here, \textsc{SpanNer} plays two roles, (1) as a base model to identify potential named entities; (2) as a meta-model (combiner) to calculate the score for each potential named entity.

Given a test span $s$ and prediction label set $\hat{\mathcal{L}}$ from $m$  base systems  ($|\hat{\mathcal{L}}| = m$).
Let $\mathcal{L}$ be NER label set where $|\mathcal{L}| = c$ and $c$ is the number of pre-defined NER classes (i.e., ``LOC, ORG, PER, MISC, O'' in \texttt{CoNLL 2003 (EN)}.)

For each $l \in \mathcal{L}$ we define $P(s, l)$  as the combined probability that span $s$ can be assigned as label $l$, which can be calculated as:
\begin{align}
    P(s, l) = \sum_{\hat{l} \in \hat{\mathcal{L}} \wedge \hat{l} = l} \mathrm{score}(s, \hat{l}),
\end{align}

\noindent
where $\mathrm{score}(\cdot)$ is defined as Eq.\ref{eq:score}. Then the final prediction label is:
\begin{align}
    \argmax_{l \in (\mathcal{L})} P(s, l),
\end{align}

Intuitively, Fig.~\ref{fig:sp-combiner} gives an example of how \textsc{SpanNer} re-recognizes the entity ``\texttt{New York}'' based on outputs from four base systems. 
As a base system, \textsc{SpanNer} predicts this span as ``LOC'', and the label will be considered as one input of the combiner model.

The prediction labels of the other three base models are ``LOC'', ``ORG'', and ``O'', respectively.
Then, as a combiner, \textsc{SpanNer} calculates the score for each predicted label.
We sum weights (scores) of the same label that are predicted by the base models and select the label that achieves the maximum score as the output of the combiner.

\section{Experiment}

\subsection{Base Systems} \label{sec:basesystem}
To make a thorough evaluation of \textsc{SpanNer} as a system combiner, as illustrated in Tab.~\ref{tab:seqmodels-res}, we first implement 10 \textsc{SeqLab} based systems that cover rich combinations of popular neural components. To be fair for other system combination methods, we also include two \textsc{SpanNer}s as base systems.
To reduce the uncertainty, we run experiments with multiple trials and also perform the significant test with Wilcoxon Signed-Rank Test~\cite{wilcoxon1970critical} at $p<0.05$.

Regarding \textsc{SeqLab}-base systems, following \cite{fu2020rethinking}, their designs are diverse in four components:
(1) character/subword-sensitive representation: \texttt{ELMo} \citep{peters2018deep}, \texttt{Flair} \citep{akbik2018contextual,akbik2019pooled}, \texttt{BERT}~\footnote{We view BERT as the subword-sensitive representation because we get the representation of each subword. } \citep{devlin2018bert} 2) word representation: \texttt{GloVe} \citep{pennington2014glove}, \texttt{fastText}~\citep{bojanowski2017enriching};
(3) sentence-level encoders: \texttt{LSTM} \citep{hochreiter1997long}, \texttt{CNN} \citep{kalchbrenner2014convolutional,chen2019grn:};
(4) decoders: \texttt{CRF} \citep{lample2016neural,collobert2011natural}. 
We keep the testing result from the model with the best performance on the development set, terminating training when the performance of the development set is not improved in 20 epochs.

\subsection{Baselines} 
\label{sec:baseline}
We extensively explore six system combination methods as competitors, which involves supervised and unsupervised fashions.

\subsubsection{Voting-based Approaches}
\label{sec:voting}
Voting, as an \textit{unsupervised method}, has been commonly used in existing works:

\noindent\textbf{Majority voting (VM):} All the individual classifiers are combined into a final system based on the majority voting.

\noindent\textbf{Weighted voting base on overall F1-score (VOF1):} The taggers are combined according to the weights, which is the overall F1-score on the testing set.

\noindent\textbf{Weighted voting base on class F1-score (VCF1):} Also weighted voting, the weights are the categories' F1-score.

\renewcommand\tabcolsep{1.1pt}
\renewcommand{\arraystretch}{1.1}
\begin{table*}[htb]
  \centering \scriptsize
    \begin{tabular}{lcccccccccccccccc}
 \toprule
    \multicolumn{1}{l}{\multirow{3}[6]{*}{Cm}} & \multirow{2}[4]{*}{Best} & \multirow{2}[4]{*}{SpNER} & \multicolumn{3}{c}{Voting-based } & \multicolumn{3}{c}{Stacking-based } & \multirow{2}[4]{*}{Best} & \multirow{2}[4]{*}{SpNER} & \multicolumn{3}{c}{Voting-based } & \multicolumn{3}{c}{Stacking-based} \\
\cmidrule(lr){4-6}\cmidrule(lr){7-9}\cmidrule{12-14}\cmidrule{15-17}           &       &       & VM$_{\dag}$    & VOF1$_{\dag}$  & VCF1$_{\dag}$  & SVM$_{\dag}$   & RF$_{\dag}$    & XGB$_{\dag}$   &       &       & VM$_{\dag}$    & VOF1$_{\dag}$  & VCF1$_{\dag}$  & SVM$_{\dag}$   & RF$_{\dag}$    & XGB$_{\dag}$ \\
\cmidrule(lr){2-9}\cmidrule(lr){10-17}         & \multicolumn{7}{c}{CoNLL-2003 (EN)}                             &       & \multicolumn{8}{c}{OntoNotes5.0-BN (BN)} \\
\cmidrule(r){1-1}\cmidrule(lr){2-9}\cmidrule(lr){10-17}
   all  & \cellcolor{mygrey}93.02 & \cellcolor{mygrey}\textbf{93.80} & \cellcolor{mygrey}93.62 & \cellcolor{mygrey}93.57 & \cellcolor{mygrey}93.60 & \cellcolor{mygrey}93.28 & \cellcolor{mygrey}93.04 & \cellcolor{mygrey}92.93 & 90.93 & \textbf{91.14} & 90.92 & 91.29 & 91.12 & 89.67 & 90.95 & 90.50 \\
    m[:10] & \cellcolor{mygrey}93.02 & \cellcolor{mygrey}\textbf{93.78} & \cellcolor{mygrey}93.48 & \cellcolor{mygrey}93.55 & \cellcolor{mygrey}93.54 & \cellcolor{mygrey}93.21 & \cellcolor{mygrey}93.03 & \cellcolor{mygrey}93.18 & 90.93 & \textbf{91.48} & 91.03 & 91.41 & 91.27 & 89.97 & 90.92 & 90.91 \\
    m[:9] & \cellcolor{mygrey}93.02 & \cellcolor{mygrey}\textbf{93.81} & \cellcolor{mygrey}93.57 & \cellcolor{mygrey}93.59 & \cellcolor{mygrey}93.51 & \cellcolor{mygrey}93.33 & \cellcolor{mygrey}93.26 & \cellcolor{mygrey}93.35 & 90.93 & \textbf{91.64} & 91.16 & 91.24 & 91.22 & 90.16 & 90.75 & 90.76 \\
    m[:8] & \cellcolor{mygrey}93.02 & \cellcolor{mygrey}\textbf{93.81} & \cellcolor{mygrey}93.41 & \cellcolor{mygrey}93.52 & \cellcolor{mygrey}93.54 & \cellcolor{mygrey}93.28 & \cellcolor{mygrey}93.17 & \cellcolor{mygrey}93.14 & 90.93 & \textbf{91.74} & 91.17 & 91.59 & 91.39 & 90.16 & 90.69 & 90.81 \\
    m[:7] & \cellcolor{mygrey}93.02 & \cellcolor{mygrey}\textbf{93.72} & \cellcolor{mygrey}93.41 & \cellcolor{mygrey}93.47 & \cellcolor{mygrey}93.41 & \cellcolor{mygrey}93.26 & \cellcolor{mygrey}92.98 & \cellcolor{mygrey}93.00 & 90.93 & \textbf{91.86} & 91.60 & 91.66 & 91.57 & 90.16 & 90.80 & 90.73 \\
    m[:6] & \cellcolor{mygrey}93.02 & \cellcolor{mygrey}\textbf{93.71} & \cellcolor{mygrey}93.21 & \cellcolor{mygrey}93.63 & \cellcolor{mygrey}93.53 & \cellcolor{mygrey}93.20 & \cellcolor{mygrey}93.27 & \cellcolor{mygrey}93.21 & 90.93 & \textbf{91.95} & 91.42 & 91.74 & 91.67 & 90.34 & 91.09 & 91.04 \\
    m[:5] & \cellcolor{mygrey}93.02 & \cellcolor{mygrey}\textbf{93.80} & \cellcolor{mygrey}93.46 & \cellcolor{mygrey}93.54 & \cellcolor{mygrey}93.52 & \cellcolor{mygrey}93.33 & \cellcolor{mygrey}93.19 & \cellcolor{mygrey}93.28 & 90.93 & 90.65 & 91.69 & 91.77 & \textbf{91.97} & 90.54 & 90.72 & 90.69 \\
    m[:4] & \cellcolor{mygrey}93.02 & \cellcolor{mygrey}\textbf{93.70} & \cellcolor{mygrey}93.29 & \cellcolor{mygrey}93.69 & \cellcolor{mygrey}93.61 & \cellcolor{mygrey}93.47 & \cellcolor{mygrey}93.20 & \cellcolor{mygrey}93.28 & 90.93 & 90.30 & \textbf{91.18} & 91.13 & 90.32 & 90.02 & 90.93 & 90.77 \\
    m[:3] & \cellcolor{mygrey}93.02 & \cellcolor{mygrey}\textbf{93.75} & \cellcolor{mygrey}93.66 & \cellcolor{mygrey}93.75 & \cellcolor{mygrey}93.61 & \cellcolor{mygrey}93.30 & \cellcolor{mygrey}93.38 & \cellcolor{mygrey}93.43 & 90.93 & \textbf{91.13} & 91.07 & \textbf{91.13} & \textbf{91.13} & 90.89 & 90.98 & 91.08 \\
    m[:2] & \cellcolor{mygrey}93.02 & \cellcolor{mygrey}93.01 & \cellcolor{mygrey}\textbf{93.02} & \cellcolor{mygrey}92.99 & \cellcolor{mygrey}92.95 & \cellcolor{mygrey}92.74 & \cellcolor{mygrey}92.86 & \cellcolor{mygrey}92.86 & 90.93 & 89.81 & \textbf{90.70} & 89.78 & 90.01 & 90.61 & 90.98 & 90.81 \\
    m[2:4] & \cellcolor{mygrey}92.41 & \cellcolor{mygrey}\textbf{93.66} & \cellcolor{mygrey}92.41 & \cellcolor{mygrey}92.46 & \cellcolor{mygrey}92.78 & \cellcolor{mygrey}92.32 & \cellcolor{mygrey}92.37 & \cellcolor{mygrey}92.51 & 90.53 & 90.23 & 88.54 & \textbf{90.53} & 89.10 & 89.38 & 90.26 & 90.18 \\
    m[4:6] & \cellcolor{mygrey}92.11 & \cellcolor{mygrey}\textbf{93.39} & \cellcolor{mygrey}92.01 & \cellcolor{mygrey}92.11 & \cellcolor{mygrey}92.31 & \cellcolor{mygrey}92.01 & \cellcolor{mygrey}92.15 & \cellcolor{mygrey}92.17 & 89.43 & \textbf{90.80} & 89.33 & 89.43 & 89.77 & 89.66 & 89.49 & 89.99 \\
    m[3:6] & \cellcolor{mygrey}92.28 & \cellcolor{mygrey}\textbf{93.04} & \cellcolor{mygrey}92.97 & \cellcolor{mygrey}92.92 & \cellcolor{mygrey}92.95 & \cellcolor{mygrey}92.18 & \cellcolor{mygrey}92.52 & \cellcolor{mygrey}92.46 & 89.66 & \textbf{90.98} & 90.82 & 90.96 & 90.90 & 89.48 & 90.59 & 90.56 \\
    m[1:] & \cellcolor{mygrey}92.46 & \cellcolor{mygrey}\textbf{93.68} & \cellcolor{mygrey}93.59 & \cellcolor{mygrey}93.54 & \cellcolor{mygrey}93.55 & \cellcolor{mygrey}93.07 & \cellcolor{mygrey}92.83 & \cellcolor{mygrey}93.00 & 90.70 & \textbf{91.21} & 90.81 & 90.94 & 90.91 & 89.50 & 90.90 & 90.56 \\
    m[2:] & \cellcolor{mygrey}92.41 & \cellcolor{mygrey}\textbf{93.58} & \cellcolor{mygrey}93.43 & \cellcolor{mygrey}93.40 & \cellcolor{mygrey}93.34 & \cellcolor{mygrey}93.06 & \cellcolor{mygrey}92.96 & \cellcolor{mygrey}92.89 & 90.53 & \textbf{90.97} & 90.54 & 90.86 & 90.74 & 89.29 & 90.72 & 90.53 \\
    m[3:] & \cellcolor{mygrey}92.28 & \cellcolor{mygrey}\textbf{93.58} & \cellcolor{mygrey}93.35 & \cellcolor{mygrey}93.41 & \cellcolor{mygrey}93.35 & \cellcolor{mygrey}93.09 & \cellcolor{mygrey}92.81 & \cellcolor{mygrey}92.81 & 89.66 & \textbf{90.71} & 90.25 & 90.38 & 90.31 & 89.05 & 90.26 & 90.10 \\
    m[4:] & \cellcolor{mygrey}92.11 & \cellcolor{mygrey}\textbf{93.50} & \cellcolor{mygrey}92.86 & \cellcolor{mygrey}93.21 & \cellcolor{mygrey}93.10 & \cellcolor{mygrey}92.88 & \cellcolor{mygrey}92.79 & \cellcolor{mygrey}92.68 & 89.43 & \textbf{90.70} & 89.46 & 89.89 & 89.84 & 89.10 & 89.20 & 89.05 \\
    m[5:] & \cellcolor{mygrey}92.01 & \cellcolor{mygrey}\textbf{93.54} & \cellcolor{mygrey}92.67 & \cellcolor{mygrey}92.84 & \cellcolor{mygrey}92.78 & \cellcolor{mygrey}92.81 & \cellcolor{mygrey}92.85 & \cellcolor{mygrey}92.65 & 89.33 & \textbf{90.39} & 89.46 & 89.42 & 89.58 & 88.61 & 89.32 & 88.93 \\
    m[6:] & \cellcolor{mygrey}91.99 & \cellcolor{mygrey}\textbf{93.32} & \cellcolor{mygrey}91.85 & \cellcolor{mygrey}92.51 & \cellcolor{mygrey}92.34 & \cellcolor{mygrey}92.16 & \cellcolor{mygrey}92.58 & \cellcolor{mygrey}92.51 & 89.21 & \textbf{89.94} & 88.51 & 89.27 & 89.08 & 88.56 & 88.75 & 88.57 \\
    m[7:] & \cellcolor{mygrey}91.57 & \cellcolor{mygrey}\textbf{92.66} & \cellcolor{mygrey}90.92 & \cellcolor{mygrey}91.55 & \cellcolor{mygrey}91.29 & \cellcolor{mygrey}91.93 & \cellcolor{mygrey}92.20 & \cellcolor{mygrey}92.02 & 89.03 & \textbf{89.42} & 88.33 & 88.33 & 88.62 & 88.00 & 87.87 & 87.86 \\
    m[8:] & \cellcolor{mygrey}90.88 & \cellcolor{mygrey}\textbf{91.29} & \cellcolor{mygrey}87.39 & \cellcolor{mygrey}90.65 & \cellcolor{mygrey}90.32 & \cellcolor{mygrey}90.98 & \cellcolor{mygrey}90.90 & \cellcolor{mygrey}90.83 & 86.84 & \textbf{88.52} & 86.50 & 87.61 & 87.19 & 86.35 & 87.14 & 87.10 \\
    m[9:] & \cellcolor{mygrey}89.71 & \cellcolor{mygrey}\textbf{91.21} & \cellcolor{mygrey}85.97 & \cellcolor{mygrey}87.31 & \cellcolor{mygrey}86.27 & \cellcolor{mygrey}89.68 & \cellcolor{mygrey}89.50 & \cellcolor{mygrey}89.56 & 86.18 & \textbf{88.36} & 85.87 & 86.27 & 86.20 & 85.34 & 86.20 & 86.01 \\
    m[10:] & \cellcolor{mygrey}83.03 & \cellcolor{mygrey}\textbf{85.65} & \cellcolor{mygrey}78.49 & \cellcolor{mygrey}83.03 & \cellcolor{mygrey}81.83 & \cellcolor{mygrey}83.06 & \cellcolor{mygrey}83.17 & \cellcolor{mygrey}83.17 & 83.87 & \textbf{86.52} & 81.05 & 83.87 & 83.39 & 82.25 & 83.94 & 83.86 \\
    \midrule
    Std.  &     \cellcolor{mygrey}--  & \cellcolor{mygrey}1.76  & \cellcolor{mygrey}3.50   & \cellcolor{mygrey}2.48  & \cellcolor{mygrey}2.78  & \cellcolor{mygrey}2.19  & \cellcolor{mygrey}2.15  & \cellcolor{mygrey}2.16  &--       & 1.28  & 2.44  & 1.95  & 2.01  & 1.97  & 1.84  & 1.85 \\
    \midrule
    Avg.  & \cellcolor{mygrey}91.98 & \cellcolor{mygrey}\textbf{93.00} & \cellcolor{mygrey}91.83 & \cellcolor{mygrey}92.36 & \cellcolor{mygrey}92.22 & \cellcolor{mygrey}92.24 & \cellcolor{mygrey}92.22 & \cellcolor{mygrey}92.21 & 89.73 & \textbf{90.45} & 89.63 & 90.02 & 89.88 & 89.00 & 89.72 & 89.63 \\
    
    \bottomrule
    \end{tabular}
    \vspace{-6pt}
        \caption{System combination results on CoNLL-2003 (EN) and OntoNotes5.0-BN (BN) datasets. \textit{SpNER} denotes \textsc{SpanNer} while \textit{all} setting denotes that all the models are putting together.
        \textit{Avg.} and \textit{Std.} represents the Average and Standard Deviation, respectively.
        $\mathbf{m}[i:k]$ is a group of models whose performance descending ranking are located on $[i,k)$. 
        \textit{Best} denotes the best performance of the single model on a model group $\mathbf{m}[i:k]$. $\dag$ shows that the combined result of the baseline is significantly worse than the \textsc{SpanNer} (with Wilcoxon Signed-Rank Test at $p<0.05$). 
        The values in bold indicate the best-combined results.
        }
  \label{tab:exp-III}%
\end{table*}%

\subsubsection{Stacking-based Approaches} \label{sec:stackingmethod}
Stacking (a.k.a, Stacked Generalization) is a general method of
using a high-level model to combine lower-level models to achieve greater predictive accuracy \cite{DBLP:conf/ijcai/TingG97}.
To make a comprehensive evaluation, we investigated three popular methods that are \textit{supervised learning}, thereby requiring additional training samples. Specifically, there are:

\noindent \textbf{Support Vector Machines (SVM)}~\cite{hearst1998support}   is a supervised machine learning algorithm, which can train quickly over large datasets. Therefore, the ensemble classifier is usually SVM.

\noindent \textbf{Random Forest (RF)}~\cite{breiman2001random} is a common ensemble classifier that randomly selects a subset of training samples and variables to make multiple decision trees. 

\noindent \textbf{Extreme Gradient Boosting (XGB)}~\cite{chen2016xgboost} is also an ensemble machine learning algorithm. It is based on the decision-tree and the gradient boosting decision~\cite{friedman2000additive}.

Notably, compared to our model, these methods are computationally expensive since they require external training samples for system combiners,  which is achieved by (i) collecting training data by performing five-fold cross-validation \cite{wu2003stacked,florian2003named}  on the original training samples of each dataset (ii) training a system combiner based on collected samples.

\subsection{Exp-III: Nuanced View}
\paragraph{Setup}
Most previous works on system combination only consider one combination case where all base systems are put together. In this setting, we aim to explore  more fine-grained combination cases.
Specifically, we first sort systems based on their performance in a descending order to get a list $\mathbf{m}$. We refer to $\mathbf{m}[i:k]$ as one combination case, dubbed \textit{combined interval}, which represents systems whose ranks are between $i$ and $k$. 
In practice, we consider 23 combination cases showing in Tab.~\ref{tab:exp-III}. 
To examine whether the \textsc{SpanNner} is significantly better than the other baseline methods, we conduct the significance test with Wilcoxon Signed-RankTest~\cite{wilcoxon1970critical} at $p<0.05$.

\paragraph{Results}
Tab.~\ref{tab:exp-III} shows results of our \textsc{SpanNer} against six baseline combiner methods on \texttt{CoNLL-2003} and \texttt{OntoNotes5.0-BN} under a nuanced view.
We can observe that: 

\noindent (1) Overall, our proposed \textsc{SpanNer}
outperforms all other competitors significantly (p-value $<$ 0.05) on most of the combination cases include the one (``\textit{all}'') that most previous works have explored.

\noindent (2) As more base systems are introduced in descending order, the combined performance will be improved gradually. 
The combination performance will decrease with the reduction of the best single system, which holds for all the combiners.

\noindent (3) The best performance is always achieved on the combination case with more models, instead of the one with a small number of top-scoring base models. This suggests that introducing more base models with diverse structures will provide richer complementary information.

\subsection{Exp-IV: Aggregated View}

\paragraph{Setup}
To also explore the effectiveness of \textsc{SpanNer} on the other datasets, we calculate the average performance of each system combination method over 23 combination cases.

\paragraph{Results}
Tab.~\ref{tab:exp-IV} shows the results, and we can observe that:
\noindent comparing with the three voting combiner, \textsc{SpanNer} achieves the best average combination performance with the lowest standard deviation, which holds for seven of nine testing datasets with statistical significance p$<$0.05.
Specifically, the performance gap between \textsc{SpanNer} and other combiners is larger on datasets from web domain: \texttt{WB} and Twitter: \texttt{W16}, \texttt{W17}.

\renewcommand\tabcolsep{2.5pt}
\begin{table}[!htb]
  \centering \footnotesize
    \begin{tabular}{lcccccccc}
    \toprule
    \multirow{2}[4]{*}{Data}  & \multicolumn{2}{c}{\textsc{SpanNer}} & \multicolumn{2}{c}{VM} & \multicolumn{2}{c}{VOF1} & \multicolumn{2}{c}{VCF1} \\
\cmidrule(lr){2-3} \cmidrule(lr){4-5}\cmidrule(lr){6-7}\cmidrule(lr){8-9}            & Avg.  & Std.  & Avg.  & Std.  & Avg.  & Std.  & Avg.  & Std. \\
    \midrule
    NW    & \cellcolor{mygrey}\textbf{90.78}$^{\dag}$ & \cellcolor{mygrey}1.1   & 90.30 & 1.4   & \cellcolor{mygrey}90.42 & \cellcolor{mygrey}1.4   & 90.44 & 1.3 \\
    BC     & \cellcolor{mygrey}\textbf{81.54}$^{\dag}$ & \cellcolor{mygrey}1.7   & 80.04 & 3.6   & \cellcolor{mygrey}80.51 & \cellcolor{mygrey}3.1   & 80.65 & 3.0 \\
    MZ     & \cellcolor{mygrey}89.17 & \cellcolor{mygrey}1.3   & 88.43 & 3.2   & \cellcolor{mygrey}88.96 & \cellcolor{mygrey}2.0   & \textbf{89.57} & 2.2 \\
    WB     & \cellcolor{mygrey}\textbf{67.45}$^{\dag}$ & \cellcolor{mygrey}2.5   & 64.57 & 5.3   & \cellcolor{mygrey}65.33 & \cellcolor{mygrey}5.0   & 66.14 & 4.6 \\
    TC     & \cellcolor{mygrey}68.25 & \cellcolor{mygrey}3.8   & 66.16 & 6.5   & \cellcolor{mygrey}67.54 & \cellcolor{mygrey}5.6   & \textbf{68.73} & 5.5 \\
    W16    & \cellcolor{mygrey}\textbf{41.60}$^{\dag}$ & \cellcolor{mygrey}6.4   & 33.23 & 9.2   & \cellcolor{mygrey}36.19 & \cellcolor{mygrey}8.9   & 39.92 & 7.9 \\
    W17    & \cellcolor{mygrey}\textbf{45.97}$^{\dag}$ & \cellcolor{mygrey}6.1   & 41.27 & 9.3   & \cellcolor{mygrey}43.32 & \cellcolor{mygrey}8.2   & 44.45 & 7.7 \\
    ES     & \cellcolor{mygrey}\textbf{87.26}$^{\dag}$ & \cellcolor{mygrey}2.6   & 86.23 & 4.3   & \cellcolor{mygrey}87.24 & \cellcolor{mygrey}2.8   & 87.00 & 2.8 \\
    NL    & \cellcolor{mygrey}\textbf{89.92}$^{\dag}$ & \cellcolor{mygrey}3.4   & 87.59 & 6.5   & \cellcolor{mygrey}88.93 & \cellcolor{mygrey}4.7   & 88.66 & 5.0 \\
    \bottomrule
    \end{tabular}
    \vspace{-6pt}
   \caption{The average system combination results on 23 combination cases of nine datasets. 
      \textit{Avg.} and \textit{Std.} denotes Average and Standard Deviation, respectively.
        $\dag$  denotes that the \textsc{SpanNer} is better than the best baseline combiner significantly (p $<$ 0.05). 
        The values in bold represent the best combination results.
        }
  \label{tab:exp-IV}
\end{table}

\subsection{Exp-VI: Interpretable Analysis}

\paragraph{Setup} 
The above experiments have shown the superior performance of \textsc{SpanNer} on system combination. To further investigate where the gains of the \textsc{SpanNer} come from, similar to \S\ref{exp-II}, we perform fine-grained evaluation on \texttt{CoNLL-2003} dataset using one combination case to interpret how  \textsc{SpanNer} outperform other (i) \textit{base systems} and (ii) other \textit{baseline combiners}.
The combination case contains base systems:  $sq0$-$5$ together with $sp1$, $sp2$ (model's detail can refer to Tab.\ref{tab:seqmodels-res}).

\paragraph{Results}
As shown in Tab.~\ref{tab:exp-VI-heatmap}, we can find:

\noindent (1) \textsc{SpanNer} v.s. \texttt{Base systems}:
the improvements of all base systems largely come from entities with  low label consistency (\texttt{eCon}: XS, S).
Particularly,  base systems with \textsc{SeqLab} benefit a lot from short entities while base systems with \textsc{SpanNer} gain mainly from long entities.

\noindent (2) \textsc{SpanNer} v.s. \texttt{Other combiners}:
as a system combiner, the improvement of \textsc{SpanNer} against other baselines mainly comes from  low label consistency (\texttt{eCon}: XS, S). By contrast, traditional combiners surpass  \textsc{SpanNer} when dealing with long sentences (\texttt{sLen}: XL).

\renewcommand{\arraystretch}{1.0}
\renewcommand\tabcolsep{2.2pt}
\begin{table}[!ht]
  \centering
    \begin{tabular}{rrrrrrrrrrr }
    \toprule
    \multicolumn{11}{c}{\textsc{SpanNer} v.s. \texttt{Base systems}} \\
    \cmidrule(lr){1-11}
     \multicolumn{4}{c}{eCon}      & \multicolumn{2}{c}{sLen} & \multicolumn{2}{c}{eLen} & \multicolumn{3}{c}{oDen} 
    \\
    \cmidrule(lr){1-11}
      \multicolumn{4}{c}{\includegraphics[scale=0.19]{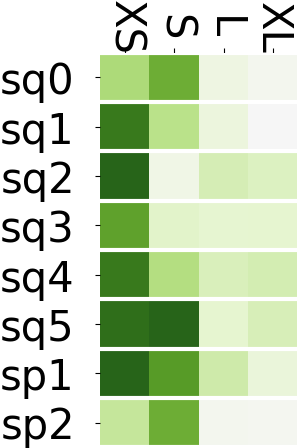}} 
        & \multicolumn{2}{c}{\includegraphics[scale=0.19]{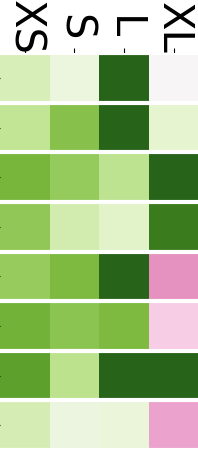}}                 
        & \multicolumn{2}{c}{\includegraphics[scale=0.19]{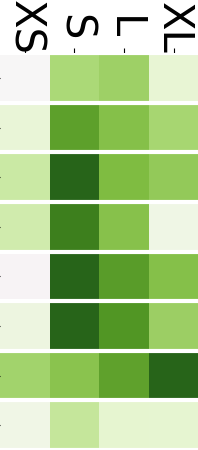}} 
        & \multicolumn{3}{c}{\includegraphics[scale=0.19]{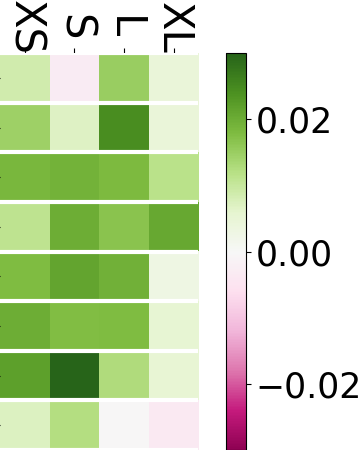}} 
        \\
    \midrule
    \midrule
    \multicolumn{11}{c}{\textsc{SpanNer} v.s. \texttt{Other combiners}} \\
     \cmidrule(lr){1-11}
     \multicolumn{4}{c}{eCon}      & \multicolumn{2}{c}{sLen} & \multicolumn{2}{c}{eLen} & \multicolumn{3}{c}{oDen} \\
     \cmidrule(lr){1-11}
           \multicolumn{4}{c}{\includegraphics[scale=0.19]{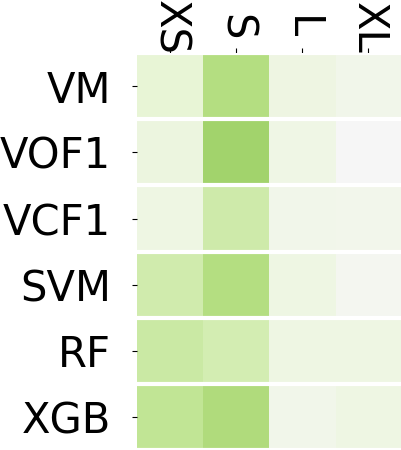}} 
        & \multicolumn{2}{c}{\includegraphics[scale=0.19]{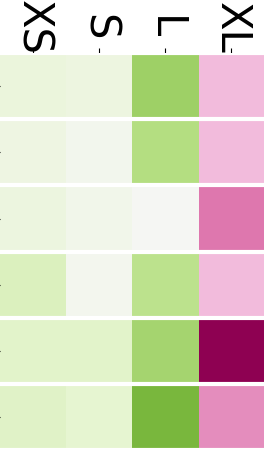}}                 
        & \multicolumn{2}{c}{\includegraphics[scale=0.19]{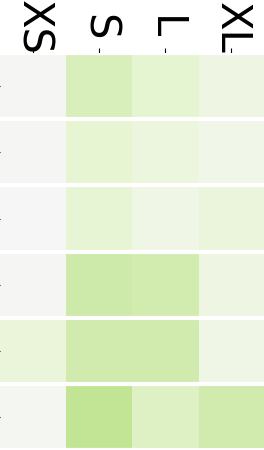}} 
        & \multicolumn{3}{c}{\includegraphics[scale=0.19]{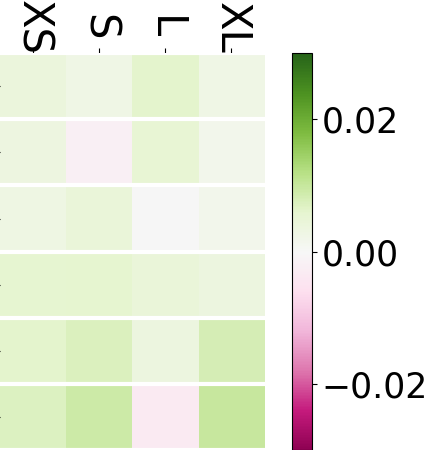}} 
        \\
  
    \bottomrule
    \end{tabular}
    \vspace{-6pt}
      \caption{Performance heatmap driven diagnosis analysis over different entity attributes. Each entry value is the \textbf{F1 difference} between the proposed \textsc{SpanNer} combiner against the \texttt{Base systems} (e.g., seq2 in Tab.\ref{tab:seqmodels-res} ) or \texttt{other combiners} (e.g.,  VM: majority voting in \S\ref{sec:voting}) respectively. The green area indicates \textsc{SpanNer} combiner perform better.}
  \label{tab:exp-VI-heatmap}
\end{table}

\section{Related Work}
\paragraph{NER as Different Tasks}
Although NER is commonly formulated as a sequence labeling task~\cite{chiu2015named,huang2015bidirectional,ma2016end,lample2016neural,akbik2018contextual,peters2018deep,devlin2018bert,xia2019multi,akbik2019pooled,luo2020hierarchical,lin2020triggerner}, recently other new forms of frameworks have been explored and have shown impressive results. For example, \citep{jiang-etal-2020-generalizing,ouchi2020instance,yu-etal-2020-named} shift NER from token-level tagging to span-level prediction task while \citep{li-etal-2020-unified,mengge2020coarse} conceptualize it as reading comprehension task.
In this work we aim to interpret the complementarity between sequence labeling and span prediction.

\paragraph{System Combination}
Traditionally, system combination was used to improve the performance of statistical MT systems~\citep{gonzalez-rubio-etal-2011-minimum, watanabe-sumita-2011-machine, duh2011generalized, mizumoto-matsumoto-2016-discriminative}. Some recent work~\citep{zhou-etal-2017-neural, DBLP:conf/ijcai/HuangZTWLXSL20} also extended this method to neural MT where the meta-model and base systems are all neural models.
There is a handful of works about system combination for NER.
\cite{wu2003stacked,florian2003named} investigated stacking and voting methods for combining strong classifiers.
\citet{ekbal2011weighted,zhang2014weighted} proposes a weighted voting approach based on differential evolution.
These works commonly require training samples and rely heavily on feature engineering.

\section{Implications and Future Directions}

\paragraph{Co-evolution of NLP Systems and their combiners}
Systems for NLP tasks (e.g., NER model) and their combiners (e.g., ensemble learning for NER) are developing in two parallel directions.
This paper builds the connection between them and proposes a model that can be utilized as both a base NER system and a system combiner.
Our work opens up a direction toward making the algorithms of NLP models and system combination co-evolved.
The unified idea can be applied to other NLP tasks, and some traditional methods like re-ranking in syntactic parsing can be \textit{re-visited}.
For example, we can formulate constituency parsing \cite{jiang-etal-2020-generalizing} as well as its re-ranking \cite{collins-koo-2005-discriminative, huang-2008-forest} as a span prediction \cite{stern2017minimal} problem, which is be unified and parameterized with the same form.

\paragraph{\text{CombinaBoard}}
It has become a trend to use a Leaderboard (e.g., paperwithcode\footnote{https://paperswithcode.com/}) to track current progress in a particular field, especially with the rapid emergence of a plethora of models.
Leaderboard makes us pay more attention to and even obsess over the state-of-the-art systems \cite{ethayarajh-jurafsky-2020-utility}.
We argue that Leaderboard with an effective system combination (dubbed \textsc{CombinaBoard}) feature would allow researchers to quickly find the complementarities among different systems. As a result, the value of a worse-ranked model still could be observed through its combined results. In this paper, we make the first step towards this by releasing a preliminary \textsc{CombinaBoard} for the NER task \url{http://spanner.sh}.
Our model also has been deployed into the \textsc{ExplainaBoard} \cite{liu2021explainaboard} platform, which allows users to flexibly perform system combination of top-scoring systems in an interactive way: \url{http://explainaboard.nlpedia.ai/leaderboard/task-ner/}

\section*{Acknowledgements}

We thank Professor Graham Neubig and anonymous reviewers for valuable feedback and helpful suggestions.
This work was supported by the Air Force Research Laboratory under agreement number FA8750-19-2-0200. The U.S. Government
is authorized to reproduce and distribute reprints for Governmental
purposes notwithstanding any copyright notation thereon. The views and
conclusions contained herein are those of the authors and should not be
interpreted as necessarily representing the official policies or
endorsements, either expressed or implied, of the Air Force Research
Laboratory or the U.S. Government.

\bibliography{anthology,acl2020}
\bibliographystyle{acl_natbib}

\appendix

\section{Attribute Interval}
The detailed attribute interval for attributes: \texttt{eCon}, \texttt{sLen}, \texttt{eLen}, and \texttt{oDen}.

\renewcommand\tabcolsep{1.6pt}
\begin{table}[htb]
  \centering \footnotesize
    \begin{tabular}{lllll}
    \toprule
    Bucket & \texttt{eCon}  & \texttt{sLen}  & \texttt{eLen}  & \texttt{oDen} \\
    \midrule
    XS    & [0.0] & [1, 7] & [1]  & [0] \\
    S     & [0, 0.5] & [7, 16] & [2]  & [0, 0.067] \\
    L     & [0.5, 0.999] & [16, 31] & [3]  & [0.067, 0.203] \\
    XL    & [1.0]  & [31, 124] & [3, 6.0] & [0.203, 1.0] \\
    \bottomrule
    \end{tabular}
  \label{tab:addlabel}
    \caption{The attribute interval of bucket XS, S, L, and XL for attributes \texttt{eCon}, \texttt{sLen}, \texttt{eLen}, and \texttt{oDen}.}
\end{table}

\renewcommand\tabcolsep{1.8pt}
\renewcommand{\arraystretch}{1.1}
\begin{table}[!thb]
  \centering \footnotesize
    \begin{tabular}{lcccccccccc }
    \toprule
    \multicolumn{1}{c}{\multirow{2}[4]{*}{Models}} & \multicolumn{5}{c}{Char/Sub.}      & \multicolumn{3}{c}{Word} & \multicolumn{2}{c}{Sent.}  \\
    \cmidrule(lr){2-6}\cmidrule(lr){7-9}\cmidrule(lr){10-11}          & \rotatebox{90}{none}  & \rotatebox{90}{cnn}   & \rotatebox{90}{elmo}  & \rotatebox{90}{flair} & \rotatebox{90}{bert}  & \rotatebox{90}{none}  & \rotatebox{90}{rand}  & \rotatebox{90}{glove} & \rotatebox{90}{lstm}  & \rotatebox{90}{cnn}     \\
    \midrule
    CflairWglove\_lstmCrf (sq0) &       &       &       & $\surd$     &       &       &       & $\surd$     & $\surd$     &       \\
     CflairWnone\_lstmCrf (sq1) &       &       &       & $\surd$     &       & $\surd$    &       &       & $\surd$     &        \\
     CbertWglove\_lstmCrf (sq2) &       &       &       &       & $\surd$     &       &       & $\surd$     & $\surd$     &       \\
      CbertWnon\_lstmCrf (sq3) &       &       &       &       & $\surd$     & $\surd$     &       &       & $\surd$     &        \\
     CelmoWglove\_lstmCrf (sq4) &       &       & $\surd$    &       &       &       &       & $\surd$     & $\surd$    &      \\
     CelmoWnone\_lstmCrf (sq5) &       &       & $\surd$     &       &       & $\surd$     &       &       & $\surd$     &       \\
     CcnnWglove\_lstmCrf (sq6) &       & $\surd$    &       &       &       &       &       & $\surd$      & $\surd$     &         \\
    CcnnWglove\_cnnCrf (sq7) &       & $\surd$     &       &       &       &       &       & $\surd$     &       & $\surd$       \\
    CcnnWrand\_lstmCrf (sp8) &       & $\surd$     &       &       &       &       & $\surd$     &       & $\surd$     &         \\
     CnoneWrand\_lstmCrf (sp9)  & $\surd$     &       &       &       &       &       & $\surd$     &       & $\surd$     &         \\
    \bottomrule
    \end{tabular}
    \vspace{-6pt}
       \caption{The illustration of \textsc{SeqLab}'s model name and its structures.
       ``Sub'', ``sent.'' and ``Dec.'' denotes the subword embedding, sentence encoder, and decoder, respectively. All the ten sequence labeling models use the CRF as the decoder. $\surd$ indicates the embedding/structure is utilized in the current \textsc{SeqLab} system. 
       }
  \label{tab:seqmodels-model-name}
\end{table}

\section{Model Name illustration of \textsc{SeqLab}}

Tab.~\ref{tab:seqmodels-model-name} illustrates the full model name and the detailed structure of the \textsc{SeqLab} models. All the \textsc{SeqLab} models use the CRF as the decoder.
For example, the full model name of ``sq0'' is ``CflairWglove\_lstmCrf'', representing a sequence labeling model that uses the Flair as character-level embedding, GloVe as word-level embedding, LSTM as the sentence-level encoder, and CRF as the decoder. For ``sq3'', its full model name is ``CbertWnon\_lstmCrf'', representing a sequence labeling model that uses the BERT as character-level embedding,  LSTM as the sentence-level encoder, and CRF as the decoder.

\end{document}